\documentclass[10pt,twocolumn,letterpaper]{article}

\usepackage[final]{cvpr}      

\usepackage{graphicx}
\usepackage{times}
\usepackage{amsmath}
\usepackage{amssymb}
\usepackage{epsfig}

\usepackage{subfigure}        

\usepackage[pagebackref,breaklinks,colorlinks]{hyperref}

\def\cvprPaperID{2361} 
\def\confName{CVPR}
\def\confYear{2022}

\begin{document}

\title{Spatial Location Constraint Prototype Loss for Open Set Recognition}

\author{Ziheng Xia,~ Ganggang Dong,~ Penghui Wang\thanks{Corresponding authors  : Hongwei Liu, Penghui Wang.},~  Hongwei Liu*\\
National Lab of Radar Signal Processing, Xidian University.\\
{\tt\small \{xiaziheng@stu., dongganggang@, wangpenghui@mail.,  hwliu@\}xidian.edu.cn.}}

\maketitle
\begin{abstract}
One of the challenges in pattern recognition is open set recognition. Compared with closed set recognition, open set recognition needs to reduce not only the empirical risk, but also the open space risk, and the reduction of these two risks corresponds to classifying the known classes and identifying the unknown classes respectively. How to reduce the open space risk is the key of open set recognition. This paper explores the origin of the open space risk by analyzing the distribution of known and unknown classes features. On this basis, the spatial location constraint prototype loss function is proposed to reduce the two risks simultaneously. Extensive experiments on multiple benchmark datasets and many visualization results indicate that our methods is superior to most existing approaches. The code is released on \url{https://github.com/Xiaziheng89/Spatial-Location-Constraint-Prototype-Loss-for-Open-Set-Recognition}.
\end{abstract}

\section{Introduction}  \label{section 1}
With the development of deep learning technology, pattern recognition based on deep neural network has been greatly developed, such as image recognition and speech recognition \cite{face-recognition,speech-recognition}. Although some deep learning methods have achieved great success, there are still many challenges and difficulties that need to be addressed in these applications. One of these challenges is that the test set may contain some classes that are not present in the training set. Unfortunately, the traditional deep neural network is incapable of detecting unknown classes. Therefore, open set recognition (\textbf{OSR}) was proposed to solve this kind of problem that the model needs to not only correctly classify known classes, but also identify unknown classes\cite{1-vs-set}. 

In contrast to OSR, the test set and the training set contain the same data categories in closed set recognition (\textbf{CSR}), and classifying known classes correctly corresponds to reducing the empirical risk. Therefore, OSR needs to reduce not only the empirical risk, but also the open space risk\cite{1-vs-set}, which corresponds to identifying unknown classes effectively.

\begin{figure*}[ht]
\centering
\subfigure[SoftMax(test set)] {\label{Softmax(test set)}     
\includegraphics[width=0.67\columnwidth]{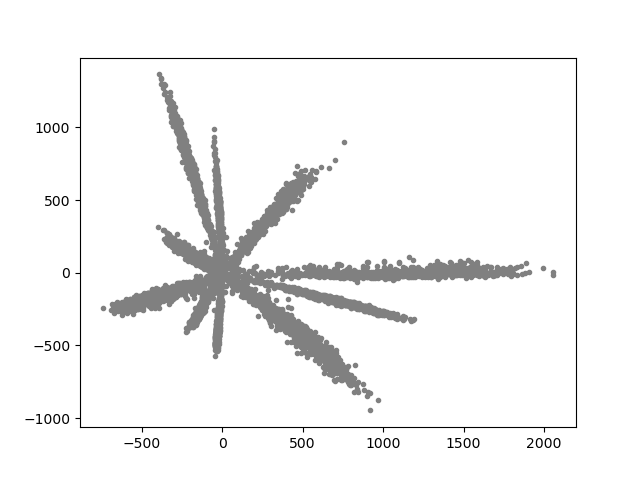}  }
\subfigure[GCPL(test set)] {\label{GCPL(test set)}     
\includegraphics[width=0.67\columnwidth]{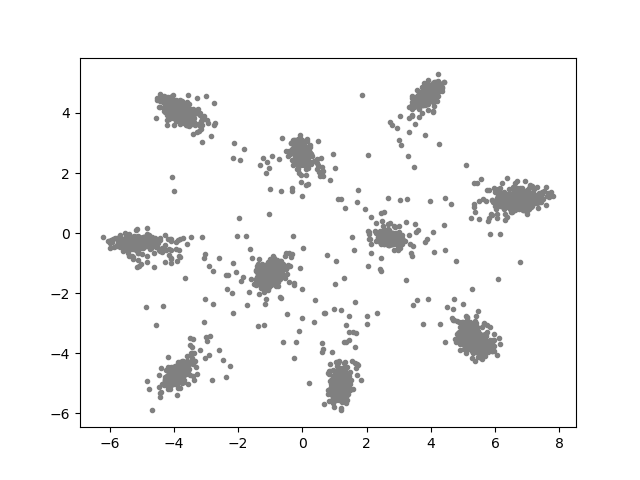}  }
\subfigure[SLCPL(test set)] { \label{SLCPL(test set)}     
\includegraphics[width=0.67\columnwidth]{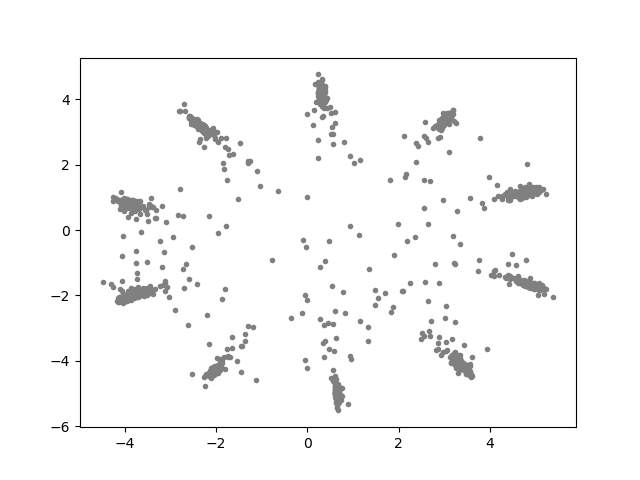}   }

\subfigure[SoftMax(open set)] {\label{Softmax(open set)}     
\includegraphics[width=0.67\columnwidth]{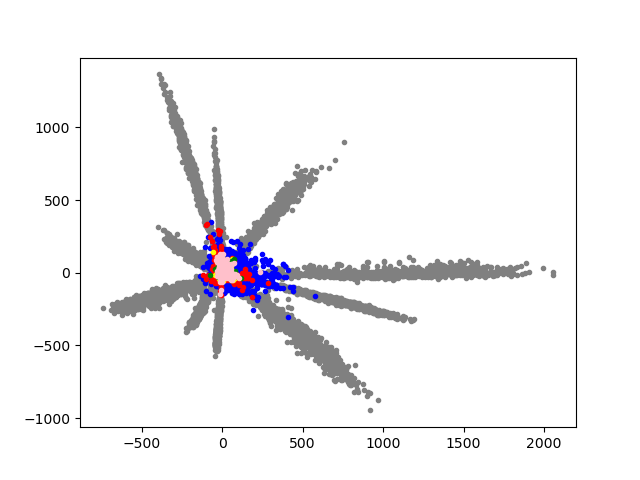}  }
\subfigure[GCPL(open set)] {\label{GCPL(open set)}     
\includegraphics[width=0.67\columnwidth]{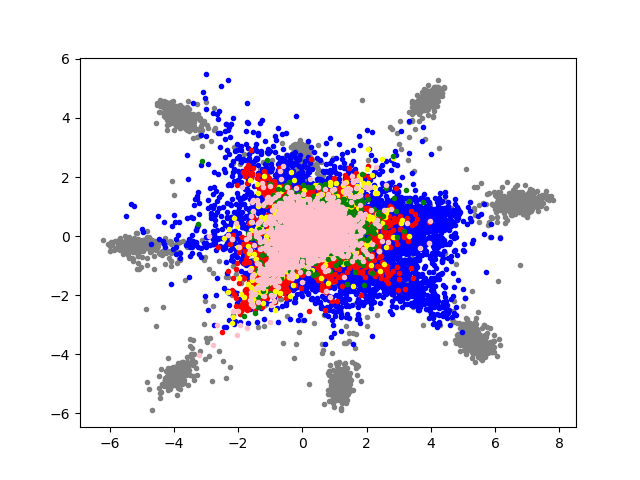}  }
\subfigure[SLCPL(open set)] { \label{SLCPL(open set)}     
\includegraphics[width=0.67\columnwidth]{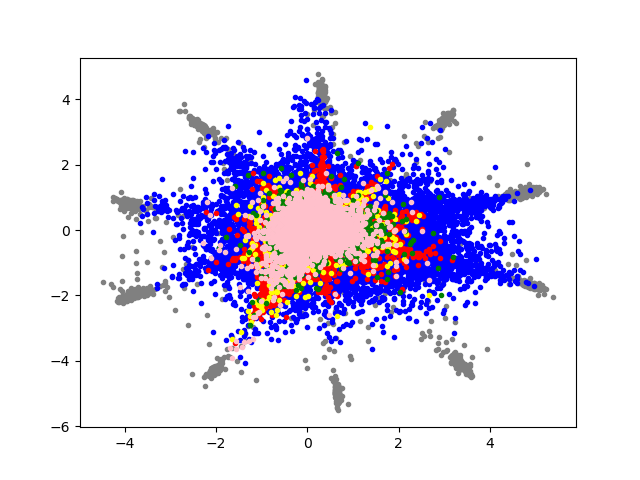}   }
\caption{\textbf{The visualization results of LeNet++ on known and unknown classes}\cite{objectosphere}. In this figure, MNIST(gray) is used for training the LeNet++ network, while KMNIST(blue), SVHN(red), CIFAR10(green), CIFAR100 (yellow) and TinyImageNet(pink) are used for open set evaluation. The first row in the figure shows the visualization results on the MNIST test set under different methods. According to the second row of the figure, our SLCPL can reduce the open space risk more effectively than other methods.}
\label{fig.1}
\end{figure*}

Due to the strong feature extraction ability, there is no doubt that deep neural network can do well in CSR. As shown in Fig.\ref{Softmax(test set)}, the LeNet++ trained with SoftMax can classify MNIST effectively. However, it can be seen from Fig.\ref{Softmax(open set)} that there is obvious overlap between known and unknown classes features. The overlapping of these features in the feature space is the direct cause of the open space risk.

In order to reduce the two risks simultaneously, Yang \etal proposed the generalized convolutional prototype learning (\textbf{GCPL}) for robust classification and OSR\cite{GCPL,CPN}. It can be seen from Fig.\ref{GCPL(test set)} that the LeNet++ trained with GCPL can classify MNIST effectively. 
With regard to the open space risk, as shown in Fig.\ref{GCPL(open set)}, although much better than SoftMax, there are still three clusters of known features that overlap with unknown features. Similar to SoftMax, it will be seen from Section \ref{section 3} that the way GCPL trains the network has disadvantages in reducing the open space risk.

As shown in Figs.\ref{Softmax(open set)} and \ref{GCPL(open set)}, we already know that the direct cause of the open space risk is that known and unknown classes features overlap in the feature space. In other words, if we can figure out why they overlap, then it's possible to significantly reduce, or even eliminate the open space risk. Therefore, the distribution of known and unknown features is further studied in this paper. And then, we find the origin of the open space risk, which is critical to address the OSR problem.

On the basis of understanding the origin of the open space risk, we propose the spatial location constraint prototype loss (\textbf{SLCPL}) for OSR, which adds a constraint term to control the spatial location of the prototypes in the feature space. As shown in Fig.\ref{SLCPL(test set)}, while effectively reducing the empirical risk, SLCPL controls the clustering of known classes in the edge region of the feature space. And it can be seen from Fig.\ref{SLCPL(open set)} that SLCPL can also reduce the open space risk effectively.

Our contributions mainly focus on the following:
\begin{itemize}
\item The distribution of known and unknown features is analyzed in detail, and then we discusse the root cause of the open space risk in this paper.
\item A novel loss function, SLCPL, is proposed to address the OSR problem.
\item Many experiments and analyses have proved our theory about the origin of the open space risk and the effectiveness of the proposed loss function.
\end{itemize}

\section{Related Work}
\subsection{Open Set Recognition}
Scheirer \etal first defined the OSR issue in 2013\cite{1-vs-set}. When deep learning performance was not as good as it is today, most of the work on OSR was based on support vector machine, such as \cite{1-vs-set,w-svm,p_i-svm,BFHC-1,BFHC-2,SVM-64,SVM-65,EPCC}. Subsequently, the traditional method represented by support vector machine is gradually replaced by the deep neural network. The open space risk is called "agnostophobia" by Dhamija \etal. Apart from using LeNet++ to assess the open space risk, they also proposed a novel Objectosphere loss function for OSR\cite{objectosphere}. Bendale \etal proposed the OpenMax model, which replaces the SoftMax layer by the OpenMax layer to predict the probability of unknown classes\cite{OpenMax}. Ge \etal combined the characteristics of generative adversarial network\cite{GAN} and OpenMax, and proposed the G-OpenMax model, which trains deep neural network with generated unknown classes\cite{G-OpenMax}. Neal \etal proposed the OSRCI model for OSR, which also adds generated samples into the training phase\cite{OSRCI}. Some works based on the reconstruction idea also make important contributions to OSR, such as \cite{CROSR}.

Although various algorithms or theories have been put forward to address the OSR problems, almost no one analyzes the root cause of the open space risk, and this paper will discuss this problem in Section \ref{section 3}.

\subsection{Prototype Learning}
The prototype is often regarded as one or more points in the feature space to represent the clustering of a specific category. Wen \etal proposed a center loss to improve the face recognition accuracy, and it can learn more discriminative features\cite{center_loss}. Similar to ~\cite{center_loss}, Yang \etal proposed the generalized convolutional prototype learning(GCPL) for robust classification\cite{GCPL}. Subsequently, this model was modified to convolutional prototype network(CPN) for OSR\cite{CPN}.

Although GCPL can effectively improve the compactness of the feature clustering, as shown in the Figs.\ref{GCPL(test set)} and \ref{GCPL(open set)}, simply increasing intra-class compactness still creates the open space risk slightly. Therefore, this paper propose a novel loss function that adds spatial location constraints to the prototype to further reduce the open space risk.

\section{The Nature of Open Set Recognition} \label{section 3}
According to the Fig.\ref{fig.1}, we believe that studying the distribution of known classes and unknown classes in the feature space is the key to solving OSR problems. Therefore, This section try to figure out why the known class features and unknown class features may overlap in the feature space.

\subsection{Feature Distribution of Known Classes}
According to the different rules of feature distribution, we divide the feature distribution of known classes into two categories: the half open space distribution, and the prototype subspace distribution.

\textbf{The half-open space distribution}. When the neural network is trained with SoftMax, the function of the last fully connected layer in the network is equivalent to the establishment of several hyper-planes in the feature space. These hyper-planes divide the feature space into several half-open spaces, with each known class occupying one corresponding half-open space. As shown in Fig.\ref{Softmax(test set)}, the $10$ known classes seem to be separated by $10$ potential rays drawn from the spatial origin, and entire $2$D feature space is divided into several half-open spaces by these potential rays.

\textbf{The prototype subspace distribution}. When the neural network is trained with the prototype learning, the prototype subspace distribution is created. In this kind of distribution, the whole feature space is divided into several known class subspace and residual space(residual space is also called the open space) by the prototypes, and each known class subspace is represented as a hyper-sphere. As shown in Fig.\ref{GCPL(test set)}, each known class feature is distributed in the circle determined by the each prototype. Because GCPL does not limit the spatial location of the prototypes, the known features are possible to be distributed in the center region of the feature space. Moreover, the feature distribution of our SLCPL also falls into this category, but our model constrains the spatial location of the prototypes to be distributed in the edge region of the feature space, as shown in Fig.\ref{SLCPL(test set)}.

In the supplementary material, we give more details and discussion about these two feature distribution.

\subsection{Feature Distribution of Unknown Classes}
From Figs.\ref{Softmax(open set)}, \ref{GCPL(open set)} and \ref{SLCPL(open set)}, we can see that whatever the unknown class is, its feature usually tend to be distributed in the center region of the entire feature space. Next, we try to explain the inevitability of this phenomenon.

As we all know, convolution neural network(\textbf{CNN}) extracts the sample features through the convolution kernel. The convolution kernels of a trained network contain template information that matches the known classes well. In the test phase, the template contained in the convolution kernel is used to match the test samples. If they match well, the convolution operation will get a higher score; Otherwise, the score will be lower. The same goes for fully connected networks.

\begin{figure}[ht]
\centering
\subfigure[The convolution kernel matches the sample.] {\label{convolution 1}     
\includegraphics[width=1.0\columnwidth]{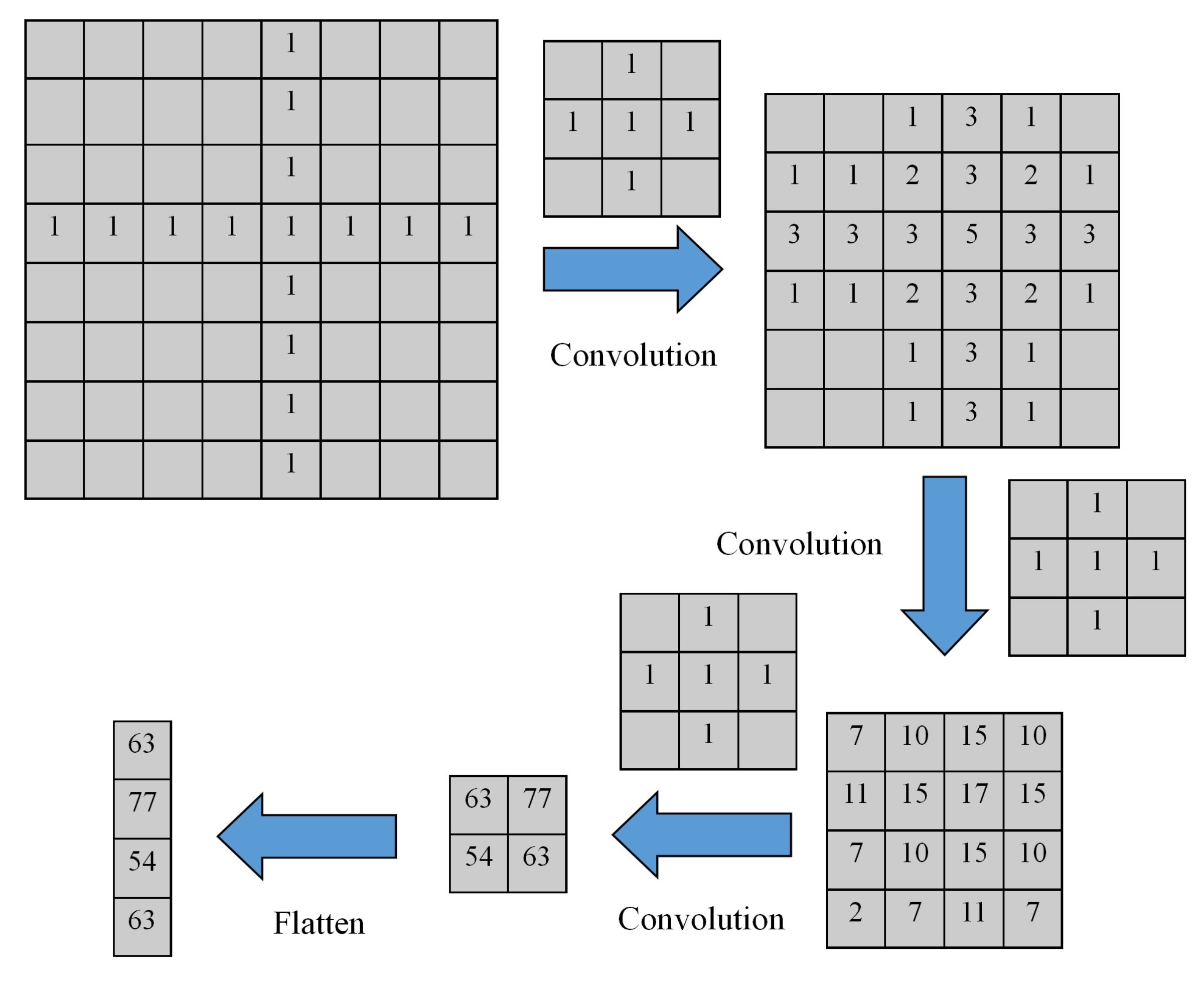}  }

\subfigure[The convolution kernel doesn't match the sample.] {\label{convolution 2}     
\includegraphics[width=1.0\columnwidth]{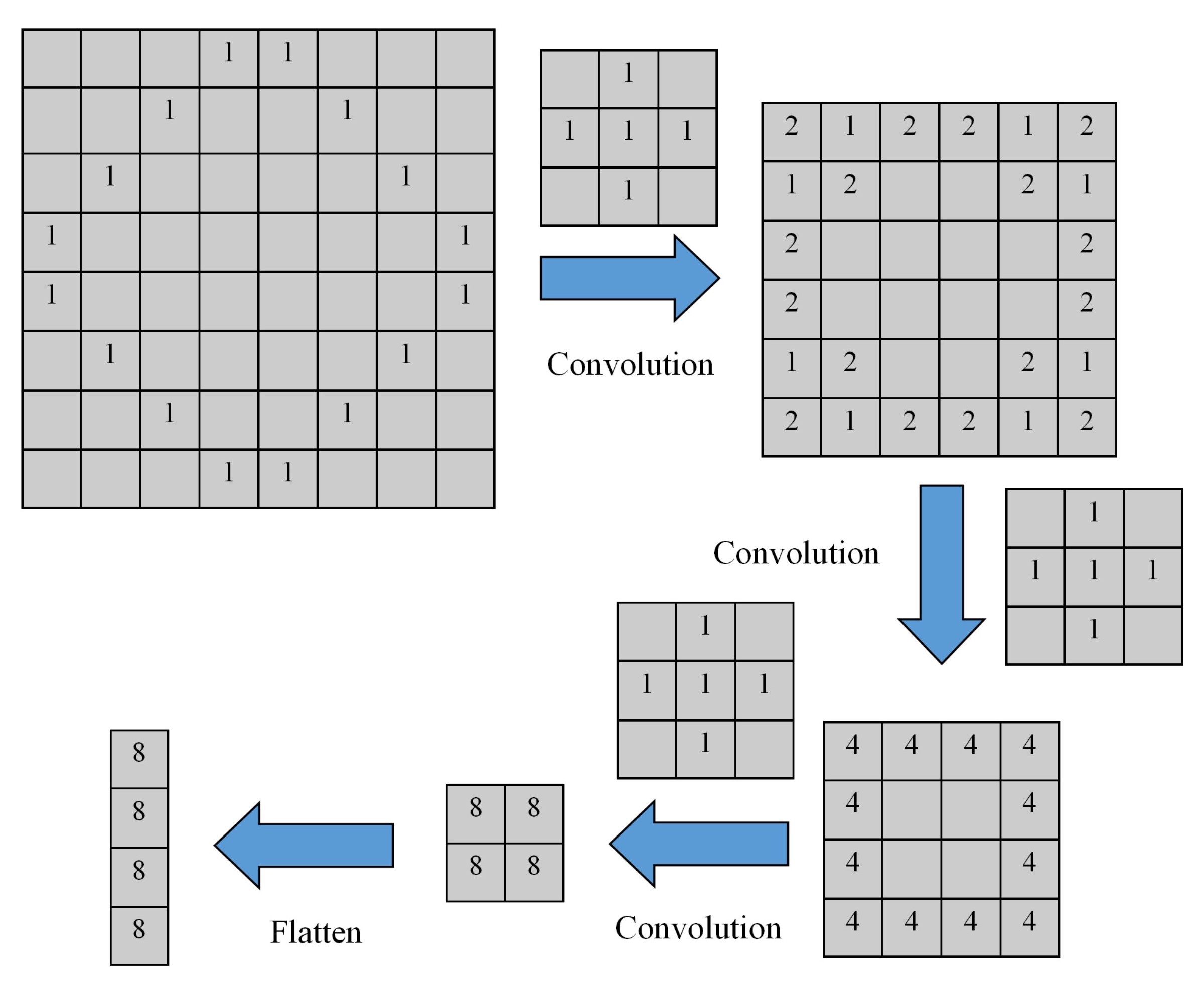}  }
\caption{\textbf{The diagram of the convolution process}.}
\label{convolution process}
\end{figure}

\begin{figure*}[ht]
\centering
\subfigure[SoftMax] {\label{softmax with 3 images}     
\includegraphics[width=0.67\columnwidth]{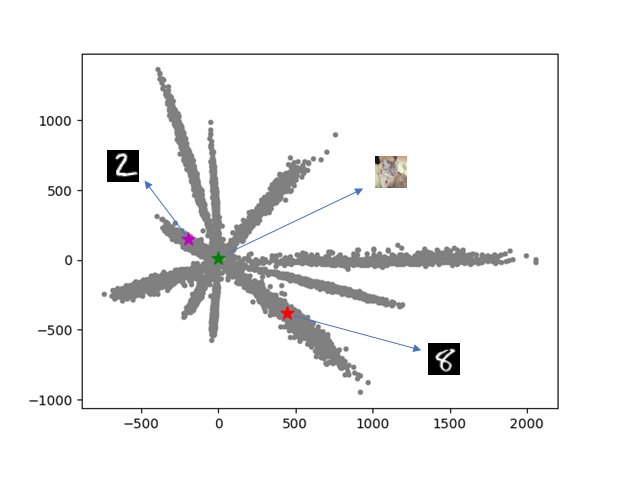}  }
\subfigure[GCPL] {\label{GCPL with 3 images}     
\includegraphics[width=0.67\columnwidth]{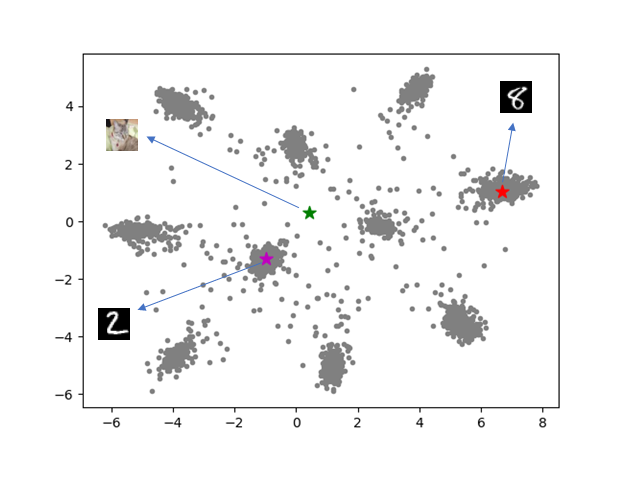}  }
\subfigure[SLCPL] {\label{SLCPL with 3 images}     
\includegraphics[width=0.67\columnwidth]{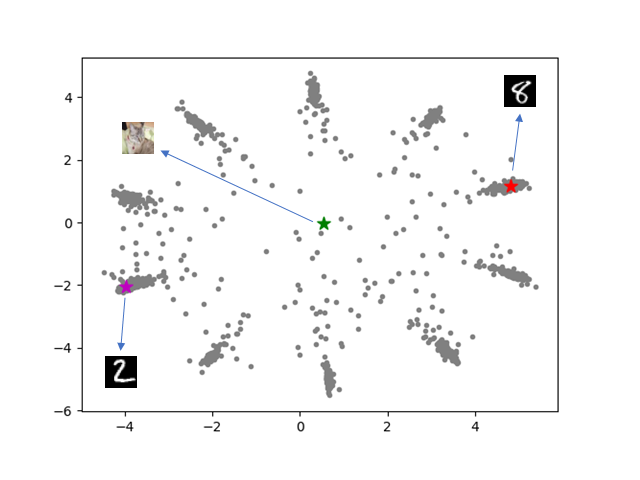}  }
\caption{\textbf{The location diagram of three specific samples in the feature space}. The stars represent the spatial location of the three samples("cat", "2" and "8").}
\label{fig.4}
\end{figure*}

\begin{figure}[ht]
\centering
\includegraphics[width=1.0\columnwidth]{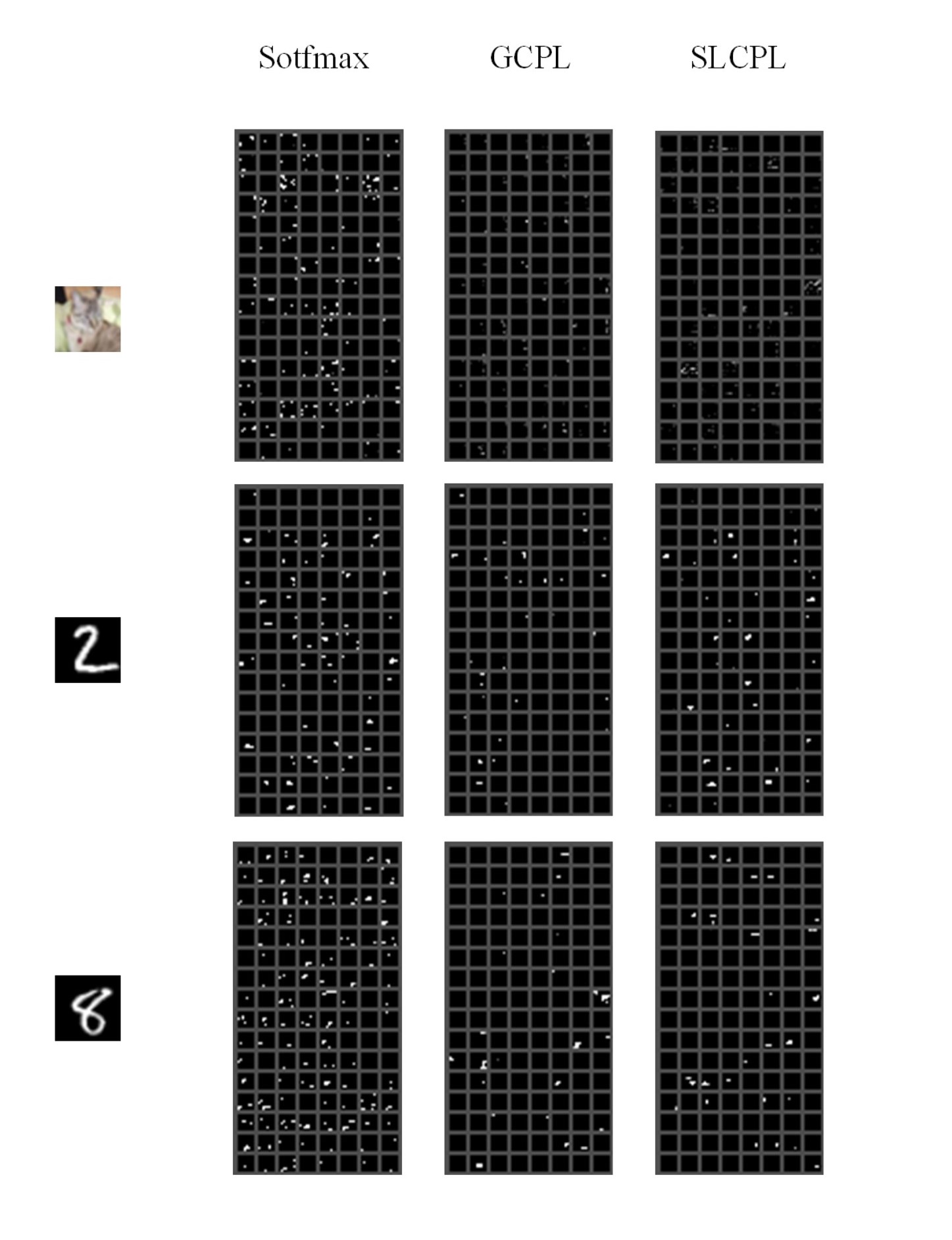}  
\caption{\textbf{The visualization results of the last convolution feature map in LeNet++}. Different $3$ columns(rows) correspond to different $3$ methods(samples). Since the last convolution feature map of the LeNet++ network is $128$ channels, each subfigure contains $128$ subgraphs.}
\label{fig.5}
\end{figure}

As shown in Fig.\ref{convolution process}, we illustrate this matching process with two simple examples of convolution. When the convolution kernel matches the sample well, it can be seen from Fig.\ref{convolution 1} that the convolution operation can get the higher value feature. These flattened features determine the position of the sample in the feature space, and higher value features correspond to higher coordinate values in the feature space. On the contrary, as shown in Fig.\ref{convolution 2}, the mismatch between the sample and the convolution kernel will result in small coordinate values of the sample in the feature space, which will make the unmatched sample distributed in the center region of the feature space.

In order to further illustrate this problem, we chose three specific samples, namely "cat", "2" and "8", and we performed visualization analysis on them. Their locations in the feature space are shown in Fig.\ref{fig.4}. Among them, the sample "cat" is located in the center region of the feature space. According to our theory, the value of the unknown feature should be less than the value of the known feature(it's strictly absolute value, but for convenience we don't make a clear specification as here). Therefore, we visualized the feature map after the last convolution operation in the LeNet++ network. As shown in Fig.\ref{fig.5}, The brighter the area, the higher the value. By comparing these $9$ subfigures, we can find the following patterns:
\begin{itemize}
\item In all three methods, the brightness of the "cat" feature is always lower than that of the samples "2" and "8".
\item The brightness of the SoftMax feature map is significantly higher than that of GCPL and SLCPL.
\item In the GCPL feature maps, the brightness of the samples "2" is slightly lower than that of the samples "8".
\item In the SLCPL feature maps, the brightness of the samples "2" is almost equal to that of the samples "8".
\end{itemize}
All these phenomena confirm the correctness of our theory. The reason why the convolution kernel is not visualized here is that it is difficult for human eyes to recognize the template contained in the convolution kernel. More details are provided in the supplementary material.

\textbf{Summary}. Finally, we summarize the origin of the open space risk: First, a CNN is trained with known classes, and most current training methods(such as SoftMax, GCPL) cannot avoid the known features distributed in the center region of the feature space. And then, because of the mismatch between unknown classes and the convolution kernel, unknown classes usually tend to be distributed in the center region of the feature space. Therefore, the open space risk occurs when known and unknown features overlap in the center region of the feature space.

\section{Spatial Location Constraint Prototype Loss}
According to the origin of the open space risk we explored in section \ref{section 3}, if we can control the known features to be distributed in the edge region of the feature space, it will reduce the open space risk effectively. Therefore, we propose the loss function SLCPL to accomplish this purpose. Since SLCPL is an improvement on GCPL, we will introduce GCPL first.

\textbf{GCPL}. Given training set $D=\{(x_1,y_1), (x_2,y_2), \cdots\}$ with $N$ known classes, the label of original data $x_i$ is $y_i \in \{1,\cdots,N\}$. A CNN is used as a classifier, whose parameters and embedding function are denoted as $\theta$ and $\Theta$. The prototypes $O=\{O^i, i=1,2,\cdots,N\}$ are initialized randomly or with a distribution(such as a Gaussian distribution). 

For a training sample $(x,y)$, the GCPL loss function can be expressed by
\begin{equation} \label{GCPL loss}
l_G(x,y;\theta,O) = l(x,y;\theta,O) + \lambda pl(x;\theta,O),
\end{equation}
where the optimization of $l(x,y;\theta,O)$ is used to classify the different known classes, $\lambda$ is a hyper-parameter, and the constraint term $pl(x;\theta,O)$  is used to make the cluster more compact.

Specifically, $l(x,y;\theta,O)$ can be expressed as
\begin{equation}
\begin{split}
l(x,y;\theta,O) &= -\log p(y=k|x,\Theta,O) \\
&= -\log\frac{e^{-d(\Theta(x), O^k)}}{\sum_{i=1}^N e^{-d(\Theta(x), O^i)}},
\end{split}
\end{equation}
where the $d(\Theta(x), O^i)$ is the Euclidean distance between $\Theta(x)$ and $O^i$. And the constraint term $pl(x;\theta,O)$ can be expressed as
\begin{equation}
pl(x;\theta,O) = \|\Theta(x^k)- O^k\|^2_2, k=1,\cdots,N,
\end{equation}
where the $x^k$ is the training samples of class $k$. Under the optimization of Eq.\eqref{GCPL loss}, the known features extracted from the network will present the prototype subspace distribution, such as Fig.\ref{GCPL(test set)}. 

\textbf{SLCPL}. According to the theory introduced in this paper, as long as the known feature clusters can be constrained in the edge region of the feature space, the open space risk can be effectively reduced. Therefore, we proposed the spatial location constraint prototype loss based on GCPL:
\begin{equation} \label{SLCPL loss}
l_S(x,y;\theta,O) = l_G(x,y;\theta,O) + slc(O).
\end{equation}

In Eq.\eqref{SLCPL loss}, the spatial location constraint term $slc(O)$ can be expressed as 
\begin{equation}
slc(O) = \frac{1}{N-1}\sum_{i=1}^N(r_i-\frac{1}{N}\sum_{j=1}^Nr_j)^2,
\end{equation}
where $r_i=d(O^i,O_c)$ and $O_c=\frac{1}{N}\sum_{i=1}^NO^i$.

In other words, the term $slc(O)$ is the variance of the distances $r_i$ between the prototypes and the center $O_c$. Here, choosing $O_c$ instead of coordinate origin as the center is more beneficial to the optimization of the training process. By controlling the variance of these distances, the known feature clusters in Fig.\ref{GCPL(test set)} can be manipulated into the distributions in Fig.\ref{SLCPL(test set)}.

\textbf{How to detect unknown classes}? In our method, known features are required to cluster. Therefore, the distance between the feature and the prototype can be used to measure the probability of which category it belongs to. Specifically, the probability that $x$ in the test set belongs to a known class can be determined by the following formula:
\begin{equation} \label{probability}
p(\hat y=k|x) \propto \ \exp\big(-\min_{k\in\{1,\cdots,N\}} d(\Theta(x), O^k)\big).
\end{equation}

Based on the distance distribution from the known features to the corresponding prototypes, a threshold value $\tau$ can be determined. When the distance between the sample feature and the prototype is greater than the threshold $\tau$, the sample will be identified as unknown class; Otherwise, the category is further determined according to Eq.\eqref{probability}. However, the optimal threshold is not easy to determine, so we provides a calibration-free measure of detection performance in the section \ref{section 5} for experiments.

\section{Experiments} \label{section 5}
\subsection{Experimental Settings} \label{section 5.1}
\textbf{Datasets}. We provide a simple summary of these protocols for each data set.
\begin{itemize}
\item \textbf{MNIST}\cite{MNIST},\textbf{SVHN}\cite{SVHN},\textbf{CIFAR10}\cite{CIFAR}.  Each of these three datasets contains $10$ classes. $6$ categories are randomly selected as known classes, and the remaining $4$ categories are unknown classes.
\item \textbf{CIFAR+10},\textbf{CIFAR+50}. For these two datasets, $4$ known classes are randomly sampled from CIFAR10 for training. $10$ and $50$ classes are randomly sampled from CIFAR100 respectively, which are used as unknown classes.
\item \textbf{TinyImageNet}\cite{TinyImageNet}. $20$ known classes and $180$ unknown classes are randomly sampled for evaluation.
\end{itemize}

\textbf{Evaluation Metrics}. We choose the accuracy and Area Under the Receiver Operating Characteristic(\textbf{AUROC}) curves to evaluate the performance of classifying known classes and identifying unknown classes respectively\cite{AUROC}.

\textbf{Network Architecture}. In order to make a fair comparison with the evaluation results of Neal \etal\cite{OSRCI}, we also adopted the same encoder network structure for experiments.

\textbf{Other Settings}. In our experiments, the hyper-parameter $\lambda$ is set to $0.1$. The momentum stochastic gradient descent (SGD-M) optimizer is used to optimize the classifier. The initial learning rate of the network is set to $0.1$, dropping to one-tenth of the original rate every $30$ epochs, and we train the network for $100$ epochs.

\subsection{Ablation Study}
An important factor affecting OSR performance is the openness, which is defined by \cite{1-vs-set}, and it can be expressed as
\begin{equation}
\mathbb{O} = 1 - \sqrt{\frac{2*N_{train}}{N_{test}+N_{target}}},
\end{equation}
where $N_{train}$ is the number of known classes, $N_{test}$ is the number of test classes that will be observed during testing, and $N_{target}$ is the number of target classes that needs to be correctly recognized during testing. 

In this section, we analyze the proposed methods on CIFAR100\cite{CIFAR}, which consists of $100$ classes. We randomly sample $15$ classes out of $100$ classes as known classes and varying the number of unknown classes from $15$ to $85$, which means openness is varied from $18\%$ to $49\%$. The performance is evaluated by the macro-average F1-score in $16$ classes($15$ known classes and unknown). For ablation study, the network model only trained with $l(x,y;\theta,O)$ is denoted as prototype loss, abbreviated as \textbf{PL}.

\begin{figure}[ht]
\centering
\includegraphics[width=1.0\columnwidth]{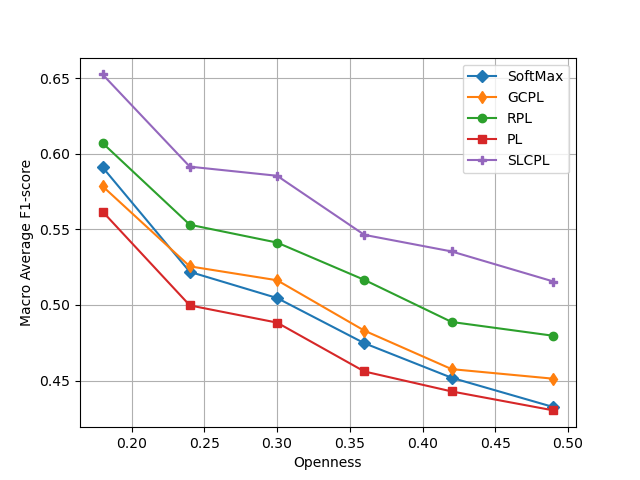}  
\caption{\textbf{Macro Average F1-score against varying Openness with different methods for ablation analysis}. }
\label{Macro Average F1-score with openness}
\end{figure}

The experimental results are shown in Fig.\ref{Macro Average F1-score with openness}. The performance of PL, GCPL, and SLPL increases in successively, which proves the effectiveness of the proposed method.

\subsection{Fundamental Experiments}
Because OSR needs to reduce the empirical risk and the open space risk simultaneously, we conducted two experiments to explore the performance of the proposed method respectively.

As shown in Tab.\ref{CSR results}, whether it is simple MNIST or complex CIFAR10, the CSR accuracy results of proposed method have an advantage in both mean and variance, which shows that our method can well reduce the empirical risk. Compared with GCPL, the constraint term $slc(O)$ proposed in this paper does not reduce the accuracy of CSR, but improves it.

\begin{table}[ht]
\begin{center}
\begin{tabular}{|c|c|c|c|}
\hline
Method & MNIST(\%) & SVHN(\%) & CIFAR10(\%) \\
\hline\hline
SoftMax\cite{OSRCI}			   & 99.5$\pm$0.2 & 94.7$\pm$0.6 & 80.1$\pm$3.2 \\ 
OpenMax\cite{OpenMax}		  & 99.5$\pm$0.2 & 94.7$\pm$0.6 & 80.1$\pm$3.2 \\
G-OpenMax\cite{G-OpenMax}   & 99.6$\pm$0.1 & 94.8$\pm$0.8 & 81.6$\pm$3.5 \\
OSRCI\cite{OSRCI}       		& 99.6$\pm$0.1 & 95.1$\pm$0.6 & 82.1$\pm$2.9 \\
CROSR\cite{CROSR}       		& 99.2$\pm$0.1 & 94.5$\pm$0.5 & 93.0$\pm$2.5 \\
RPL\cite{RPL} 						& \textbf{99.8}$\pm$0.1  & 96.9$\pm$0.4 & 94.5$\pm$1.7 \\
GCPL\cite{CPN}          	& 99.7$\pm$0.1 & 96.7$\pm$0.4 & 92.9$\pm$1.2 \\
\hline\hline
SLCPL         & \textbf{99.8}$\pm$0.1 & \textbf{97.1}$\pm$0.3 & \textbf{94.6}$\pm$1.2 \\
\hline
\end{tabular}
\end{center}
\caption{The closed set accuracy results test on various methods and datasets. Every value is averaged among five randomized trials. We report the corresponding results from \cite{CPN,OSRCI,CROSR}, and reproduce the RPL results from \cite{RPL}.}
\label{CSR results}
\end{table}

\begin{table*}
\begin{center}
\begin{tabular}{|c|c|c|c|c|c|c|}
\hline
Method & MNIST(\%) & SVHN(\%) & CIFAR10(\%) & CIFAR+10(\%) & CIFAR+50(\%) & TinyImageNet(\%) \\
\hline\hline
SoftMax\cite{OSRCI}  &97.8$\pm$0.6&88.6$\pm$1.4&67.7$\pm$3.8&81.6$\pm$-&80.5$\pm$-&57.7$\pm$-\\
OpenMax\cite{OpenMax}  &98.1$\pm$0.5&89.4$\pm$1.3&69.5$\pm$4.4&81.7$\pm$-&79.6$\pm$-&57.6$\pm$-\\
G-OpenMax\cite{G-OpenMax}&98.4$\pm$0.5&89.6$\pm$1.7&67.5$\pm$4.4&82.7$\pm$-&81.9$\pm$-&58.0$\pm$-\\
OSRCI\cite{OSRCI}    &98.8$\pm$0.4&91.0$\pm$1.0&69.9$\pm$3.8&83.8$\pm$-&82.7$\pm$-&58.6$\pm$-\\
CROSR\cite{CROSR}    &99.1$\pm$0.4&89.9$\pm$1.8 & -    & -    & -    &58.9$\pm$-\\
RPL\cite{RPL}        &99.3$\pm$-&95.1$\pm$-&\textbf{86.1}$\pm$-& 85.6$\pm$- & 85.0$\pm$- & 70.2$\pm$-\\
GCPL\cite{CPN}       &99.0$\pm$0.2&92.6$\pm$0.6&82.8$\pm$2.1&88.1$\pm$-&87.9$\pm$-&63.9$\pm$-\\
\hline\hline
SLCPL& \textbf{99.4}$\pm$0.1&\textbf{95.2}$\pm$0.8&\textbf{86.1}$\pm$1.4&\textbf{91.6}$\pm$1.7&\textbf{88.8}$\pm$0.7&\textbf{74.9}$\pm$1.4\\
\hline
\end{tabular}
\end{center}
\caption{The AUROC results of the OSR test on various methods and datasets. Every value is averaged among five randomized trials. We report the corresponding results from \cite{CPN,OSRCI,CROSR,RPL}. Standard deviation values for state of the art are not available for CIFAR+10, CIFAR+50 and TinyImageNet.}
\label{AUROC results}
\end{table*}

\begin{figure*}[ht]
\centering
\subfigure[t-SNE (MNIST, Plain CNN)] {\label{t-SNE-MNIST}     
\includegraphics[width=0.67\columnwidth]{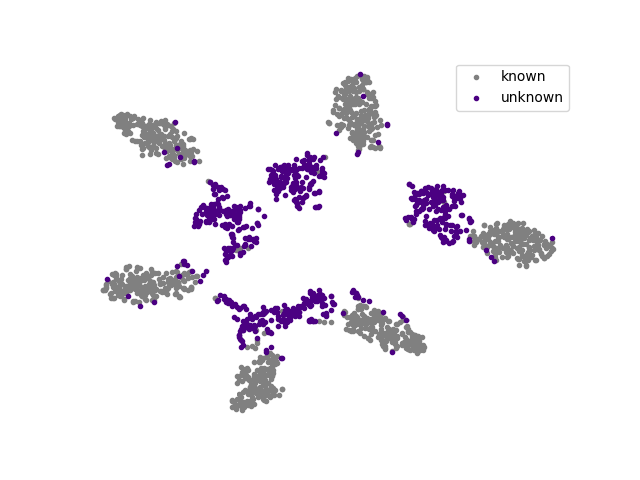}  }
\subfigure[t-SNE (open set, Plain CNN)] {\label{t-SNE-open-set}     
\includegraphics[width=0.67\columnwidth]{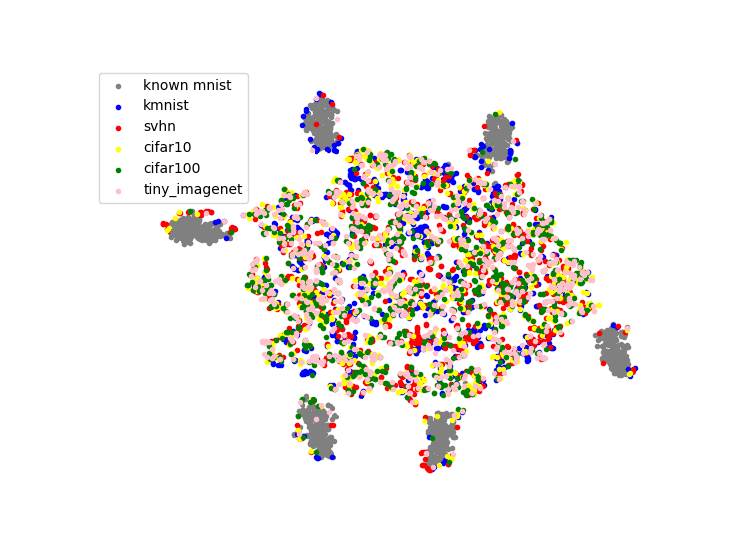}  }
\subfigure[Distance distribution diagram (Plain CNN)] {  \label{distance_1}
\includegraphics[width=0.67\columnwidth]{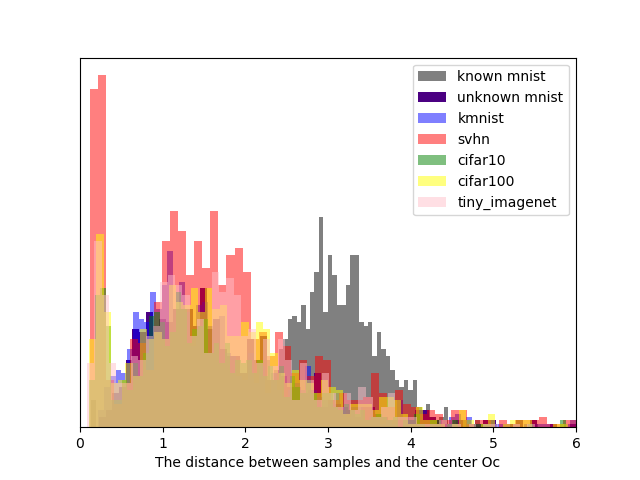}  }

\subfigure[2D-features (open set, LeNet++)] { \label{2D-open-set}     
\includegraphics[width=0.67\columnwidth]{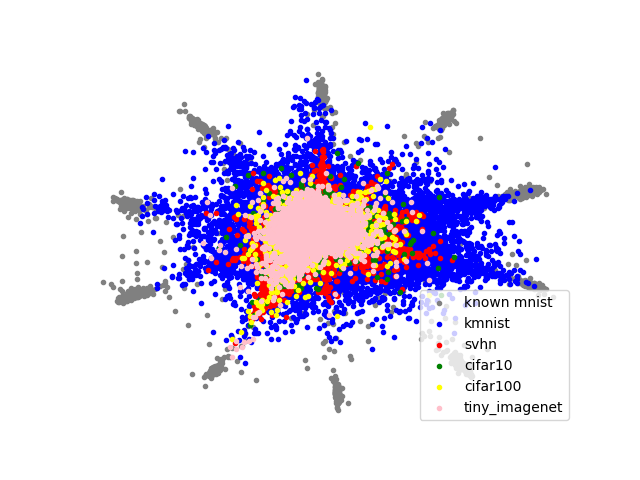}   }
\subfigure[Distance distribution diagram (LeNet++)] {    \label{distance_2}
\includegraphics[width=0.67\columnwidth]{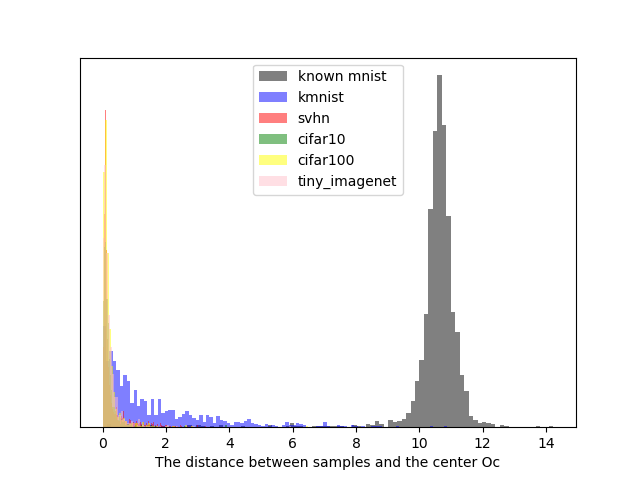}    }
\caption{\textbf{The visualization and distance distribution diagram of two network structures}. The first row of this figure corresponds the experiment results of Plain CNN in section \ref{section 5}, and the dimension of Plain CNN's feature space is $128$, so we use t-SNE to visualize its features. And the second row of this figure corresponds the experiment results of LeNet++ in section \ref{section 1}.}
\label{fig.6}
\end{figure*}

\begin{table*}[ht]
\begin{center}
\begin{tabular}{|c|c|c|c|c|}
\hline
Method          &ImageNet-crop(\%) &ImageNet-resize(\%) &LSUN-crop(\%) &LSUN-resize(\%) \\
\hline\hline
SoftMax\cite{OSRCI}             &63.9 &65.3 &64.2 &64.7 \\ 
OpenMax\cite{OpenMax}             &66.0 &68.4 &65.7 &66.8 \\
LadderNet+SoftMax\cite{CROSR}   &64.0 &64.6 &64.4 &64.7 \\
LadderNet+OpenMax\cite{CROSR}   &65.3 &67.0 &65.2 &65.9 \\
DHRNet+Softmax\cite{CROSR}      &64.5 &64.9 &65.0 &64.9 \\
DHRNet+Openmax\cite{CROSR}      &65.5 &67.5 &65.6 &66.4 \\
OSRCI\cite{OSRCI}               &63.6 &63.5 &65.0 &64.8 \\ 
CROSR\cite{CROSR}               &72.1 &73.5 &72.0 &74.9 \\
C2AE\cite{CGDL}                 &83.7 &82.6 &78.3 &80.1 \\
CGDL\cite{CGDL}                 &84.0 &83.2 &80.6 &81.2 \\
RPL\cite{RPL}                   &84.6 &83.5 &85.1 &87.4 \\
GCPL\cite{CPN}                  &85.0 &83.5 &85.3 &88.4 \\
\hline\hline
SLCPL           &\textbf{86.7} &\textbf{85.9} &\textbf{86.5} &\textbf{89.2} \\
\hline
\end{tabular}
\end{center}
\caption{Open set recognition results on CIFAR10 with various outliers added to the test set as unknowns. The performance is evaluated by macro-averaged F1-scores in $11$ classes ($10$ known classes and $1$ unknown class). We report the experiment results from \cite{CROSR,CGDL}, and reproduce the GCPL and RPL from \cite{CPN,RPL}.}
\label{ImageNet LSUN results}
\end{table*}

\begin{table}[ht]
\begin{center}
\begin{tabular}{|c|c|c|c|}
\hline
Method          & Omniglot(\%)   & M-noise(\%) & Noise(\%) \\
\hline\hline
SoftMax\cite{OSRCI}         &59.5 &80.1 &82.9 \\ 
OpenMax\cite{OpenMax}         &78.0 &81.6 &82.6 \\
CROSR\cite{CROSR}           &79.3 &82.7 &82.6 \\
CGDL\cite{CGDL}             &85.0 &88.7 &85.9 \\
RPL\cite{RPL}               &96.4 &92.6 &99.1 \\
GCPL\cite{CPN}              &97.1 &93.5 &99.5 \\
\hline\hline
SLCPL           &\textbf{97.9} &\textbf{93.7} &\textbf{99.6} \\
\hline
\end{tabular}
\end{center}
\caption{Open set recognition results on MNIST with various outliers added to the test set as unknowns. The rest of the information is similar to Tab.\ref{ImageNet LSUN results}.}
\label{Omniglot results}
\end{table}

Similar to Tab.\ref{CSR results}, as the difficulty of OSR increases, it can be seen from Tab.\ref{AUROC results} that the performance difference between different methods becomes larger and larger. In this experiment, the method proposed in this paper has the strongest ability to identify unknown classes on all data sets. In particular, the comparison with GCPL results further proves the validity of the constraint term $slc(O)$ proposed in this paper. Compared with GCPL, only adding a constraint term to the loss function can greatly improve the OSR performance, which is derived from the analysis of the origin of the open space risk.

\subsection{Visualization and Distance Analysis}
In order to better illustrate the theory and model presented in this paper, this section presents the experiment results on SLCPL through visualization and distance analysis. For convenience, the network described in section \ref{section 5.1} is denoted here as Plain CNN.

As shown in the first row of Fig.\ref{fig.6}, in this experiment, known classes come from $6$ categories in MNIST, while the remaining $4$ categories in MNIST are used as unknown classes, and this data category setting can be seen from the $10$ clusters in Fig.\ref{t-SNE-MNIST}. In the Fig.\ref{t-SNE-open-set}, KMNIST, SVHN, CIFAR10, CIFAR100 and TinyImageNet are used for open set evaluation, and it can be seen from this figure that the network can distinguish the known and unknown classes(open set), except that a few open set data features overlap with known features. In the Fig.\ref{distance_1}, we evaluated the distance distribution of known and unknown features to the center $O_c$, which corresponds to Fig.\ref{t-SNE-open-set}. As shown in Fig.\ref{distance_1}, most of the unknown features have smaller distances to the spatial center than the known features.

In the second row of Fig.\ref{fig.6}, we show the experimental results when the network structure is LeNet++. The distance distribution in Fig.\ref{distance_2} corresponds to the feature distribution in Fig.\ref{2D-open-set}. As shown in Fig.\ref{distance_2}, almost all of the unknown features are less distant from the center of feature space than the known features.

The distance analysis results of the two different network structures in Figs.\ref{distance_1} and \ref{distance_2} both prove the view of this paper that unknown features usually tend to be distributed in the center region of the feature space. In addition, although the results of these $2$ figures prove our theory, it is obvious that the performance in Fig.\ref{distance_2} is better than that in Fig.\ref{distance_1}, which may be related to the network architecture and the dimension size of the feature space.

\subsection{Further Experiments}
\textbf{Experiment I}. All samples from the $10$ classes in CIFAR10 are considered as known classes, and samples from ImageNet and LSUN are selected as unknown samples\cite{LSUN}. The Settings for data sets are the same as in \cite{CROSR}, and the other experimental settings are the same as in Section \ref{section 5.1}. The results are shown in Tab.\ref{ImageNet LSUN results}. 

\textbf{Experiment II}. All samples from the $10$ classes in MNIST are considered as known classes, and samples from Omniglot\cite{Omniglot}, MNIST-noise(denoted as M-noise) and Noise are selected as unknown samples. Omniglot is a data set containing various alphabet characters. Noise is a synthesized data set by setting each pixel value independently from a uniform distribution on $[0, 1]$. M-Noise is also a synthesized data set by adding noise on MNIST testing samples. The Settings for data sets are the same as in \cite{CROSR}, and the other experimental settings are the same as in Section \ref{section 5.1}. The results are shown in Tab.\ref{Omniglot results}. 

In these $2$ experiments, we can see from the results that on all given datasets, the proposed method is more effective than previous methods and achieves a new state-of-the-art performance. Especially, compared with GCPL, these results once again prove the validity of the constraint term $slc(O)$ proposed in this paper.

\section{Conclusion}
As mentioned in this paper, the difficulty of OSR lies in how to reduce the open space risk. To reduce the open space risk, we need to understand why it occurs first. Therefore, this paper explores the direct and root causes of the open space risk. As discussed in section \ref{section 3}, the distribution of known and unknown features determines the open space risk. Because of the mismatch between the unknown class samples and the network convolution kernel, the unknown features usually tend to be distributed in the center region of the feature space. Therefore, it is natural for us to think that if the known features can be distributed in the edge region of the feature space by adjusting the loss function, then the open space risk can be effectively reduced. Based on the above analysis and discussion, we propose the SLCPL loss function. Finally, a large number of experiments and analysis have proved our theory of the open space risk and the effectiveness of our SLCPL.

{\small
\bibliographystyle{unsrt}
\bibliography{ref_2nd}
}

\end{document}